\documentclass[12pt]{article}
\usepackage{fullpage}

\pdfoutput = 1

\usepackage{natbib}
\usepackage{setspace}
\usepackage{authblk}

\usepackage[utf8]{inputenc} 
\usepackage[T1]{fontenc}    
\usepackage{hyperref}       
\usepackage{url}            
\usepackage{booktabs}       
\usepackage{amsfonts}       
\usepackage{nicefrac}       
\usepackage{microtype}      
\usepackage{xcolor}         

\usepackage[pdftex]{graphicx}
\usepackage{subcaption}

\usepackage{amsfonts}       
\usepackage{amssymb}
\usepackage{amsmath}
\usepackage{amsthm}
\usepackage{mathtools}
\usepackage{latexsym,bm}
\DeclareMathOperator{\E}{\mathbb{E}}
\DeclareMathOperator{\R}{\mathbb{R}}
\usepackage{multicol}
\usepackage{multirow}

\usepackage{color}

\usepackage{array}
\usepackage{algorithm}
\usepackage{algorithmic}

\usepackage[shortlabels]{enumitem}

\begin{document}

\title{Diffusion Models for Time Series Applications: A Survey}
\author{Lequan Lin, Zhengkun Li, Ruikun Li, Xuliang Li, and Junbin Gao}
\affil{Discipline of Business Analytics, The University of Sydney Business School \\ The University of Sydney, Camperdown, NSW 2006, Australia\\ \{lequan.lin, zhengkun.li, ruikun.li, xuliang.li, junbin.gao\}@sydney.edu.au}
\date{}
\maketitle

\abstract{Diffusion models, a family of generative models based on deep learning, have become increasingly prominent in cutting-edge machine learning research. With a distinguished performance in generating samples that resemble the observed data, diffusion models are widely used in image, video, and text synthesis nowadays. In recent years, the concept of diffusion has been extended to time series applications, and many powerful models have been developed. Considering the deficiency of a methodical summary and discourse on these models, we provide this survey as an elementary resource for new researchers in this area and also an inspiration to motivate future research. For better understanding, we include an introduction about the basics of diffusion models. Except for this, we primarily focus on diffusion-based methods for time series forecasting, imputation, and generation, and present them respectively in three individual sections. We also compare different methods for the same application and highlight their connections if applicable. Lastly, we conclude the common limitation of diffusion-based methods and highlight potential future research directions.
}

\section{Introduction}

Diffusion models, a family of deep learning-based generative models, have risen to prominence in the machine learning community in recent years  \citep{croitoru2022diffusion, yang2022diffusion}. With exceptional performance in various real-world applications such as image synthesis \citep{austin2021structured,  dhariwal2021diffusion, ho2022cascaded}, video generation \citep{harvey2022flexible,ho2022video,yang2022video}, natural language processing \citep{li2022diffusion,savinov2022step, yu2022latent}, and time series prediction \citep{rasul2021autoregressive, li2022generative, lopezalcaraz2023diffusionbased}, diffusion models have demonstrated their power over many existing generative techniques. 

Given some observed data $\bm{x}$ from a target distribution $q(\bm{x})$, the objective of a generative model is to learn a generative process that produces new samples from $q(\bm{x})$ 
\citep{luo2022understanding}. To learn such a generative process, most diffusion models begin with progressively disturbing the observed data by injecting Gaussian noises, then applying a reversed process with a learnable transition kernel to recover the data \citep{sohl2015deep, ho2020denoising,luo2022understanding}. Typical diffusion models assume that after a certain number of noise injection steps, the observed data will become standard Gaussian noises. So, if we can find the probabilistic process that recovers the original data from standard Gaussian noises, then we can generate similar samples using the same probabilistic process with any random standard Gaussian noises as the starting point. 

The recent three years have witnessed the extension of diffusion models to time series-related applications, including time series forecasting \citep{rasul2021autoregressive, li2022generative, bilovs2022modeling}, time series imputation\citep{tashiro2021csdi, lopezalcaraz2023diffusionbased, liu2023pristi}, and time series generation \citep{lim2023regular}. Given observed historical time series, we often try to predict future time series. This process is known as time series forecasting. Since observed time series are sometimes incomplete due to reasons such as data collection failures and human errors, time series imputation is implemented to fill in the missing values. Different from time series forecasting and imputation, time series generation or synthesis aims to produce more time series samples with similar characteristics as the observed period. 

Basically, diffusion-based methods for time series applications are developed from three fundamental formulations, including denoising diffusion probabilistic models (DDPMs), score-based generative models (SGMs), and stochastic differential equations (SDEs). The target distributions learned by the diffusion components in different methods often involve the condition on previous time steps. Nevertheless, the design of the diffusion and denoising processes varies with different objectives of different tasks. Hence, a comprehensive and self-contained summary of relevant literature will be an inspiring beacon for new researchers who just enter this new-born area and experienced researchers who seek for future directions. Accordingly, this survey aims to summarize existing literature, compare different approaches, and identify potential limitations.  

In this paper, we will review on diffusion-based models for time series applications (please refer to \textbf{Table 1} for a quick summary). For a better understanding, we will include a brief introduction about three predominant formulations of diffusion models in section \ref{basics}. Next, we will categorize the existing models based on their major functions. More specifically, we will discuss the models primarily for time series forecasting, time series imputation, and time series generation in section \ref{ts_forecasting}, section \ref{ts_imputation}, and section \ref{ts_generation}, respectively. In each section, we will have a separate subsection for problem formulation, which helps to clarify the objective, training and forecasting settings of each specific application. We will highlight if a model can serve multiple purposes and articulate the linkage when one model is related to or slightly different from another. Eventually, we will conclude this survey in section \ref{conclusion}.

\begin{table*}[t!] 
\centering
\scriptsize
\renewcommand\arraystretch{1.2}
\caption{A summary of diffusion-based methods for time series applications.}
\begin{tabular}{c|c|c|cl}
\cline{1-4}
Application                                                                        & Data Type                                                                            & Method                 & Source                                                 &  \\ \cline{1-4}
\multirow{6}{*}{\begin{tabular}[c]{@{}c@{}}Time Series Forecasting\end{tabular}} & \multirow{4}{*}{\begin{tabular}[c]{@{}c@{}}Multivariate Time Series\end{tabular}} & TimeGrad               & \cite{rasul2021autoregressive}        &  \\ \cline{3-4}
                                                                                    &                                                                                      & ScoreGrad              & \cite{yan2021scoregrad}               &  \\ \cline{3-4}
                                                                                    &                                                                                      & $\text{D}^3\text{VAE}$ & \cite{li2022generative}               &  \\ \cline{3-4}
                                                                                    &                                                                                      & DSPD/CSPD              & \cite{bilovs2022modeling}             &  \\ \cline{2-4}
                                                                                    & \multirow{2}{*}{\begin{tabular}[c]{@{}c@{}}Spatio-temporal Graphs\end{tabular}}   & DiffSTG                & \cite{wen2023diffstg}                 &  \\ \cline{3-4}
                                                                                    &                                                                                      & GCRDD                  & \cite{ruikun2023}                     &  \\ \cline{1-4}
\multirow{4}{*}{\begin{tabular}[c]{@{}c@{}}Time Series Imputation\end{tabular}}  & \multirow{3}{*}{\begin{tabular}[c]{@{}c@{}}Multivariate Time Series\end{tabular}}  & CSDI                   & \cite{tashiro2021csdi}                &  \\ \cline{3-4}
                                                                                    &                                                                                      & DSPD/CSPD              & \cite{bilovs2022modeling}             &  \\ \cline{3-4}
                                                                                    &                                                                                      & SSSD                   & \cite{lopezalcaraz2023diffusionbased} &  \\ \cline{2-4}
                                                                                    & \begin{tabular}[c]{@{}c@{}}Spatio-temporal Graphs\end{tabular}                    & PriSTI                 & \cite{liu2023pristi}                  &  \\ \cline{1-4}
\begin{tabular}[c]{@{}c@{}}Time Series Generation\end{tabular}                   & \begin{tabular}[c]{@{}c@{}}Multivariate Time Series\end{tabular}                  & TSGM                   & \cite{lim2023regular}                 &  \\ \cline{1-4}
\end{tabular}
\end{table*}

\section{Basics of Diffusion Models} \label{basics}
The underlying principle of diffusion models is to progressively perturb the observed data with a forward diffusion process, then recover the original data through a backward reverse process. The forward process involves multiple steps of noise injection, where the noise level changes at each step. The backward process, on the contrary, consists of multiple denoising steps that aim to remove the injected noises gradually. Normally, the backward process is parameterized by a neural network. Once the backward process has been learned, it can generate new samples from almost arbitrary initial data. Stemming from this basic idea, diffusion models are predominantly formulated in three ways: denoising diffusion probabilistic models (DDPMs), score-based generative models (SGMs), and stochastic differential equations (SDEs). In this section, we will briefly review on these three formulations of diffusion models.

\subsection{Denoising Diffusion Probabilistic Models}
DDPMs implement the forward and backward processes through two Markov chains \citep{sohl2015deep, ho2020denoising}. Let the original observed data be $\bm{x}^0$, where $0$ indicates that the data are free from the noises injected in the diffusion process. 

The forward Markov chain transforms $\bm{x}^0$ to a sequence of disturbed data $\bm{x}^1, \bm{x}^2, ..., \bm{x}^K$ with a diffusion transition kernel:
\begin{equation}
    q(\bm{x}^k|\bm{x}^{k-1}) = \mathcal{N}\left( \bm{x}^k; \sqrt{\alpha_k} \bm{x}^{k-1}, (1-\alpha_k) \bm{I} \right),
\end{equation}
where $\alpha_k \in (0,1)$ for $k = 1,2, ..., K$ are hyperparameters indicating the changing variance of the noise level at each step, and $\mathcal{N}\left(\bm{x}; \bm{\mu}, \bm{\Sigma} \right)$ is the general notation for the Gaussian distribution of $\bm{x}$ with the mean $\bm{\mu}$ and the covariance $\bm{\Sigma}$, 
respectively. 
A nice property of this Gaussian transition kernel is that we may obtain $\bm{x}^k$ directly from $\bm{x}^0$ by
\begin{equation}\label{eq_forwardproperty}
    q(\bm{x}^k|\bm{x}^{0}) = \mathcal{N}\left( \bm{x}^k; \sqrt{\Tilde{\alpha_k}} \bm{x}^{0}, (1-\Tilde{\alpha_k}) \bm{I} \right),
\end{equation}
where $\Tilde{\alpha}_k \coloneqq \prod_{i=1}^k {\alpha_i}$. Therefore, $\bm{x}^k = \sqrt{\Tilde{\alpha}_k} \bm{x}^0 + \sqrt{1-\Tilde{\alpha}_k} \bm{\epsilon}$ with $\bm{\epsilon} \sim \mathcal{N}(\bm{0}, \bm{I})$. Normally, we design $\Tilde{\alpha}_K \approx 0$ such that $q(\bm{x}^K) \coloneqq \int q(\bm{x}^K|\bm{x}^0) q(\bm{x}^0) \mathrm{d}\bm{x}^0 \approx \mathcal{N}(\bm{x}^K; 0, \bm{I})$, 
which means the starting point of the backward chain can be any standard Gaussian noises. 

The reverse transition kernel is modelled by a parameterized neural network
\begin{equation} \label{eq_reverse_kernel}
    p_{\bm{\theta}}(\bm{x}^{k-1}|\bm{x}^{k}) = \mathcal{N}\left( \bm{x}^{k-1};\bm{\mu}_{\bm{\theta}}(\bm{x}^k,k), \bm{\Sigma}_{\bm{\theta}}(\bm{x}^k,k) \right),
\end{equation}
where $\bm{\theta}$ denotes learnable parameters. Now, the remaining problem is how to estimate $\bm{\theta}$. Basically, the objective is to maximize the likelihood objective function so that the probability of observing the training sample $\bm{x}^0$ estimated by $p_{\bm{\theta}}(\bm{x}^0)$ is maximized.
This task is accomplished by minimizing the variational lower bound of the estimated negative log-likelihood $\E[-\log{p_{\bm{\theta}}(\bm{x}^0})]$ 
, that is,
\begin{equation}
    \E_{q(\bm{x}^{0:K})} \left[ -\log{p(\bm{x}^K)} - \sum_{k=1}^K \log{\frac{p_{\bm{\theta}}(\bm{x}^{k-1}|\bm{x}^k)}{q(\bm{x}^k|\bm{x}^{k-1})}} \right]
\end{equation}
where $\bm{x}^{0:K}$ denotes the sequence $\bm{x}^0,...,\bm{x}^K$. 

\cite{ho2020denoising} proposed that we could simplify the covariance matrix $\bm{\Sigma}_{\bm{\theta}} (\bm{x}^k,k)$ in Equation (\ref{eq_reverse_kernel}) as a constant-dependent matrix $\sigma_k^2\bm{I}$, where $\sigma_k^2$ controls the noise level and may vary at different diffusion steps. Besides, they rewrote the mean as a function of a learnable noise term as:
\begin{equation}
    \bm{\mu}_{\bm{\theta}}(\bm{x}^k,k) = \frac{1}{\sqrt{\alpha_k}} \left( \bm{x}^k - \zeta(k) \bm{\epsilon}_{\bm{\theta}} (\bm{x}^{k}, k) \right),
\end{equation}
where $\zeta(k)= \frac{1- \alpha_{k}}{\sqrt{1-\Tilde{\alpha}_{k}}}$, and $\bm{\epsilon}_{\bm{\theta}}$ is a noise-matching network that predicts $\bm{\epsilon}$ corresponding to inputs $\bm{x}^k$ and $k$. With the property in Equation (\ref{eq_forwardproperty}), \cite{ho2020denoising} further simplifies the objective function to 
\begin{equation}\label{eq_DDPMloss2}
    \E_{k, \bm{x}^0, \bm{\epsilon}} \left[ \delta(k) \left\|\bm{\epsilon} - \bm{\epsilon}_{\bm{\theta}} \left(\sqrt{\Tilde{\alpha}_k} \bm{x}^0 + \sqrt{1-\Tilde{\alpha}_k}\bm{\epsilon}, k \right)\right\|^2 \right],
\end{equation}
where $\delta(k) = \frac{(1-\alpha_k)^2}{2 \sigma_k^2 \alpha_k (1 - \Tilde{\alpha}_k)}$ is a positive-valued weight that can be discarded to produce better performance in practice.

Eventually, samples are generated by eliminating the noises in $\bm{x}^K \sim \mathcal{N}(\bm{x}^K; 0, \bm{I})$. More specifically, for $k = K-1, K-2, ..., 0$, 
\begin{equation*}
    \bm{x}^k \leftarrow \frac{\left( \bm{x}^{k+1} -\zeta(k+1)\bm{\epsilon}_{\bm{\theta}} (\bm{x}^{k+1}, k+1) \right) }{\sqrt{\alpha_{k+1}}} + \sigma_k \bm{z},
\end{equation*}
where $\bm{z} \sim \mathcal{N}(\bm{0},\bm{I})$ for $k = K-1, ..., 1$, and $\bm{z} = 0$ for $k = 0$.

\subsection{Score-based Generative Models}
Score-based generative models (SGMs) consist of two modules, including score matching and annealed Langevin dynamics (ALD). ALD is a sampling algorithm that generates samples with an iterative process by applying Langevin Monte Carlo at each update step \citep{song2019generative}. Stein score is an essential component of ALD. The Stein score of a density function $q(\bm{x})$ is defined as $\nabla_{\bm{x}} \log{q (\bm{x})}$. Since the true probabilistic distribution $q(\bm{x})$ is usually unknown, score matching \citep{hyvarinen2005estimation} is implemented to approximate the Stein score with a score-matching network. Here we primarily focus on denoising score matching \citep{Vincent2011} as it is empirically more efficient, but other methods such as sliced score matching \citep{song2020sliced} are also commonly mentioned in the relevant literature.

The underlying principle of denoising score matching is to process the observed data with the forward transition kernel $q(\bm{x}^k|\bm{x}^{0}) = \mathcal{N} (\bm{x}^k; \bm{x}^0, \sigma^2_k \bm{I})$, with $\sigma^2_k$ being a set of increasing noise levels for $k = 1, ..., K$, and then jointly estimate the Stein scores for the noise density distributions $q_{\sigma_1} (\bm{x}), q_{\sigma_2} (\bm{x}), ..., q_{\sigma_k} (\bm{x})$ \citep{song2019generative}. The Stein score for noise density function $q_{\sigma_k} (\bm{x})$ is defined as $\nabla_{\bm{x}} \log{q_{\sigma_k} (\bm{x})}$.

Then, the Stein score is approximated by a neural network $\bm{s}_{\bm{\theta}}(\bm{x}, \sigma_k)$, where $\bm{\theta}$ contains learnable parameters. Accordingly, the initial objective function is given as
\begin{equation}
    \E_{k,\bm{x}^0, \bm{x}^k} \left[\left\| \bm{s}_{\bm{\theta}} (\bm{x}^k,k) - \nabla_{\bm{x}^k} \log q_{\sigma_k}(\bm{x}^k)\right\| \right].
\end{equation}
With the Gaussian assumption of the forward transition kernel, a tractable version of the objective function can be found as 
\begin{equation}
    \E_{k, \bm{x}^0, \bm{x}^k} \left[ \delta(k) \left\|\bm{s}_{\bm{\theta}}(\bm{x}^k, \sigma_k) + \frac{\bm{x}^k-\bm{x}^0}{\sigma_k^2}\right\|^2 \right],
\end{equation}
where $\delta(k)$ is a positive-valued weight depending on the noise scale $\sigma_k$. 

After the score-matching network $\bm{s}_{\bm{\theta}}$ is learned, the ALD algorithm will be implemented for sampling. The algorithm is initialized with a sequence of increasing noise levels $\sigma_1, ..., \sigma_K$ and a starting point $\bm{x}^{K,0} \sim \mathcal{N}(\bm{0},\bm{I})$. For $k = K, K-1, ..., 0$, $\bm{x}^k$ will be updated with $N$ iterations that compute
\begin{align*}
    \bm{z} &\leftarrow \mathcal{N}(\bm{0},\bm{I})\\
    \bm{x}^{k,n} &\leftarrow \bm{x}^{k,n-1} + \frac{1}{2} \eta_k \bm{s}_{\bm{\theta}} \left( \bm{x}^{k,n-1}, \sigma_k \right) + \sqrt{\eta_k} \bm{z},
\end{align*}
where $n = 1, ..., N$, $\bm{z} \sim \mathcal{N}(\bm{0}, \bm{I})$, and $\eta_k$ represents the step of update. Note that after each $N$ iterations, the last output $\bm{x}^{k,N}$ will be assigned as the starting point of the next $N$ iterations, that is, $\bm{x}^{k-1,1}$. $\bm{x}^{0,N}$ will be the final sample. The role of $\bm{z}$ in this sampling process is to slightly add uncertainty such that the algorithm will not end up with almost identical samples.

\subsection{Stochastic Differential Equations}\label{sde}
DDPMs and SGMs implement the forward pass as a discrete process, which means we should carefully design the diffusion steps. To overcome this limitation, one may consider the diffusion process as continuous such that it becomes the solution of a stochastic differential equation (SDE)\citep{song2020score}. This formulation can be thought of as a generalization of the previous two formulations since both DDPMs and SGMs are discrete forms of SDEs. The backward process is modelled as a time-reverse SDE, and samples can be generated by solving this time-reverse SDE. Let $\bm{w}$ and $\Tilde{\bm{w}}$ be a standard Wiener process and its time-reverse version, respectively, and consider a continuous diffusion time $k \in [0,K]$. A general expression of SDE is 
\begin{equation}\label{eq_sde}
    \mathrm{d}\bm{x} = f(\bm{x},k)\mathrm{d}k + g (k) \mathrm{d}\bm{w},
\end{equation}
and the time-reverse SDE, as shown by \cite{anderson1982reverse}, is
\begin{equation}
    \mathrm{d}\bm{x} =  \left[ f(\bm{x},k) - g(k)^2 \nabla_{\bm{x}} \log{q_k(\bm{x})} \right]\mathrm{d}k + g(k)\mathrm{d}\bm{\Tilde{w}},
\end{equation}
In addition, \cite{song2020score} has illustrated that sampling from the probability flow ordinary differential equation (ODE) as following has the same distribution as the time-reverse SDE:
\begin{equation}
    \mathrm{d}\bm{x} = \left[ f(\bm{x},k) - \frac{1}{2} g(k)^2 \nabla_{\bm{x}} \log{q_k(\bm{x})} \right] \mathrm{d}k.
\end{equation}
Here $f(\bm{x},k)$ and $g(k)$ separately compute the drift coefficient and the diffusion coefficient for the diffusion process. $\nabla_{\bm{x}} \log{q_k(\bm{x})}$ is the Stein score corresponding to the marginal distribution of $\bm{x}^k$, which is unknown but can be learned with a similar method as in SGMs with the objective function
\begin{equation}\label{eq_sdeloss}
    \E_{k,\bm{x}^0,\bm{x}^k}\left[ \delta(k) \left\| \bm{s}_{\bm{\theta}} (\bm{x}^k,k) - \nabla_{\bm{x}^k} \log{q_{0k}(\bm{x}^k|\bm{x}^0)} \right\|^2 \right].
\end{equation}

Now, how to write the diffusion processes of DDPMs and SGMs as SDEs? Recall that $\alpha_k$ is a defined parameter in DDPMs and $\sigma^2_k$ denotes the noise level in SGMs. The SDE corresponding to DDPMs is known as variance preserving (VP) SDE, defined as
\begin{equation}\label{sde_ddpm}
    \mathrm{d}\bm{x} = -\frac{1}{2} \alpha(k) \bm{x} \mathrm{d}k + \sqrt{\alpha (k)} \mathrm{d} \bm{w},
\end{equation}
where $\alpha(\cdot)$ is a continuous function, and $\alpha \left( \frac{k}{K} \right) = K(1 - \alpha_k)$ as $K \to \infty$. For the forward pass of SGMs, the associated SDE is known as variance exploding (VE) SDE, defined as
\begin{equation}\label{sde_sgm}
    \mathrm{d}\bm{x} = \sqrt{\frac{\mathrm{d}\left[ \sigma(k)^2 \right]}{\mathrm{d}k}} \mathrm{d}\bm{w},
\end{equation}
where $\sigma(\cdot)$ is a continuous function, and $\sigma(\frac{k}{K}) = \sigma_k$ as $K \to \infty$ \citep{song2020score}. Inspired by VP SDE, \cite{song2020score} designed another SDE called sub-VP SDE that performs especially well on likelihoods, given by
\begin{equation}
    \mathrm{d}\bm{x} = -\frac{1}{2}\alpha(k) \bm{x} \mathrm{d} k + \sqrt{\alpha(k)\left(1- e^{-2 \int_0^k \alpha(s)\mathrm{d}s}\right)}\mathrm{d}\bm{w}.
\end{equation}
The objective function involves a perturbation distribution $q_{0k}(\bm{x}^k|\bm{x}^0)$ that varies for different SDEs. For the three aforementioned SDEs, their corresponding perturbation distributions are derived as
\begin{gather}
\scalebox{1}{$\begin{aligned}
    q_{0k}(\bm{x}^k|\bm{x}^0) = \left\{
    \begin{array}{ll}
        \mathcal{N}(x^k; x^0, [\sigma(k)^2-\sigma(0)^2]\bm{I}), & \text{(VP SDE)} \\
        \mathcal{N}(x^k; x^0 e^{-\frac{1}{2} \int_0^k \alpha(s)\mathrm{d}s}, [1-e^{-\int_0^k \alpha(s)\mathrm{d}s}]\bm{I}), & \text{(VE SDE)} \\
        \mathcal{N}(x^k; x^0 e^{-\frac{1}{2} \int_0^k \alpha(s)\mathrm{d}s},  [1-e^{-\int_0^k \alpha(s)\mathrm{d}s}]^2\bm{I}), & \text{(sub-VP SDE)} 
    \end{array}
\right.
\end{aligned}$}
\end{gather}
After successfully learning $\bm{s}_{\bm{\theta}} (\bm{x}, k)$, samples are produced by deriving the solutions to the time-reverse SDE or the probability flow ODE with techniques such as ALD. 

\section{Time Series Forecasting} \label{ts_forecasting}
Multivariate time series forecasting is a crucial area of study in machine learning research, with wide-ranging applications across a variety of industries. Different from univariate time series, which only track one feature over time, multivariate time series involve the historical observations of multiple features that interact with each other and evolve with time. Consequently, they provide a more comprehensive understanding of complex systems and realize more reliable predictions of future trends and behaviours. 

In recent years, generative models have been implemented for multivariate time series forecasting tasks. For example, WaveNet is a generative model with dilated causal convolutions that encode long-term dependencies for sequence prediction \citep{oord2016wavenet}. As another example, \cite{rasul2020multi} model multivariate time series with an autoregressive deep learning model, in which the data distribution is expressed by a conditional normalizing flow. Nevertheless, the common shortcoming of these models is that the functional structure of their target distributions are strictly constrained. Diffusion-based methods, on the other hand, can provide a less restrictive solution. In this section, we will discuss four diffusion-based approaches. We also include two models designed specifically for spatio-temporal graphs (i.e., spatially related entities with multivariate time series) to highlight the extension of diffusion theories to more complicated problem settings. Since relevant literature mostly focuses on multivariate time series forecasting, ``forecasting'' refers to multivariate time series forecasting in the rest of this survey unless otherwise stated.

\subsection{Problem Formulation}
Consider a multivariate time series $\bm{X}^0=\{\bm{x}^0_1, \bm{x}^0_2,...,\bm{x}^0_T|\bm{x}^0_i \in \R^D\}$, where $0$ indicates that the data is free from the perturbation in the diffusion process. The forecasting task is to predict $\bm{X}^0_p = \{\bm{x}^0_{t_0}, \bm{x}^0_{t_0+1}, ..., \bm{x}^0_{T}\}$ given the historical information $\bm{X}^0_c = \{\bm{x}^0_{1}, \bm{x}^0_{2}, ..., \bm{x}^0_{t_0-1}\}$. $\bm{X}_c^0$ is known as the context window, while $\bm{X}_p^0$ is known as the prediction interval. In diffusion-based models, the problem is formulated as learning the joint probabilistic distribution of data in the prediction interval:
\begin{equation}\label{eq_forecasting_dist}
    q\left(\bm{x}_{t_0:T}^0|\bm{x}_{1:t_0-1}^0\right) = \prod_{t=t_0}^T q\left( \bm{x}_t^0|\bm{x}_{1:t_0-1}^0 \right).
\end{equation}
Some literature also considers the role of covariates in forecasting, such as \citep{rasul2021autoregressive} and \citep{yan2021scoregrad}. Covariates are additional information that may impact the behaviour of variables over time, such as seasonal fluctuations and weather changes. Incorporating covariates in forecasting often helps to strengthen the identification of factors that drive temporal trends and patterns in data. The forecasting problem with covariates is formulated as
\begin{equation} \label{forecasting_dist_c}
    q\left(\bm{x}_{t_0:T}^0|\bm{x}_{1:t_0-1}^0,\bm{c}_{1:T}\right) = \prod_{t=t_0}^T q\left( \bm{x}_t^0|\bm{x}_{1:t_0-1}^0,\bm{c}_{1:T} \right),
\end{equation}
where $\bm{c}_{1:T}$ denotes the covariates for all time points and is assumed to be known for the whole period. 

For the purpose of training, one may randomly sample the context window followed by the prediction window from the complete training data. This process can be seen as applying a moving window with size $T$ on the whole timeline. Then, the optimization of the objective function can be conducted with the samples. Forecasting future time series is usually achieved by the generation process corresponding to the diffusion models.

\subsection{TimeGrad}\label{timegrad}
The first noticeable work on diffusion-based forecasting is TimeGrad proposed by \cite{rasul2021autoregressive}. Developed from DDPM models, TimeGrad firstly injects noises to data at each predictive time point, and then gradually denoise through a backward transition kernel conditioned on historical time series. To encode historical information, TimeGrad approximate the conditional distribution in Equation (\ref{forecasting_dist_c}) by
\begin{equation}
    \prod_{t=t_0}^T p_{\bm{\theta}} (\bm{x}_t^0|\bm{h}_{t-1}),
\end{equation}
where
\begin{equation}
    \bm{h}_t = \mathrm{RNN}_{\bm{\theta}} (\bm{x}_t^0, \bm{c}_t, \bm{h}_{t-1})
\end{equation}
is the hidden state calculated with a RNN module such as LSTM \citep{hochreiter1997long} or GRU \citep{chung2014empirical} that can preserve historical temporal information, and $\bm{\theta}$ contains learnable parameters for the overall conditional distribution and its RNN component.

The objective function of TimeGrad is in the form of a negative log-likelihood, given as
\begin{equation}
    \sum_{t=t_0}^T - \log p_{\bm{\theta}} (\bm{x}_t^0|\bm{h}_{t-1}),
\end{equation}
where for each $t \in [t_0,T]$, $- \log p_{\bm{\theta}} (\bm{x}_t^0|\bm{h}_{t-1})$ is upper bounded by
\begin{gather}
\scalebox{1}{$\begin{aligned}\label{timegrad_obj}
    \E_{k, \bm{x}_t^0, \bm{\epsilon}} \left[ \delta(k) \left\|\bm{\epsilon} - \bm{\epsilon}_{\bm{\theta}} \left(\sqrt{\Tilde{\alpha}_k} \bm{x}_t^0 + \sqrt{1-\Tilde{\alpha}_k}\bm{\epsilon},\bm{h}_{t-1}, k \right)\right\|^2 \right].
\end{aligned}$}
\end{gather}
The context window is used to generate the hidden state $\bm{h}_{t_0-1}$ for the starting point of the training process. It is not hard to see that Equation (\ref{timegrad_obj}) is very similar to Equation (\ref{eq_DDPMloss2}) except for the inclusion of hidden states to represent the historical information.

In the training process, the parameter $\bm{\theta}$ is estimated by minimizing the negative log-likelihood objective function with stochastic sampling. Then, future time series are generated in a step-by-step manner. Suppose that the last time point of the complete time series is $\Tilde{T}$. The first step is to derive the hidden state $\bm{h}_{\Tilde{T}}$ based on the last available context window. Next, the observation for the next time point $\Tilde{T}+1$ is predicted in a similar way as DDPM:
\begin{gather*}
\scalebox{1}{$\begin{aligned}
    \bm{x}^k_{\Tilde{T}+1} \leftarrow \frac{\left( \bm{x}^{k+1}_{\Tilde{T}+1} -\zeta(k+1)\bm{\epsilon}_{\bm{\theta}} (\bm{x}^{k+1}_{\Tilde{T}+1}, \bm{h}_{\Tilde{T}}, k+1) \right) }{\sqrt{\alpha_{k+1}}} + \sigma_{k+1} \bm{z},
\end{aligned}$}
\end{gather*}
The predicted $\bm{x}^k_{\Tilde{T}+1}$ should be fed back to the RNN module to obtain $\bm{h}_{\Tilde{T}+1}$ before the prediction for the next time point. The sampling process will be repeated until the desired length of the future time series is reached.

\subsection{ScoreGrad}
ScoreGrad shares the same target distribution as TimeGrad, but it is alternatively built upon SDEs, extending the diffusion process from discrete to continuous and replacing the number of diffusion steps with an interval of integration \citep{yan2021scoregrad}. ScoreGrad is composed of a feature extraction module and a conditional SDE-based score-matching module. The feature extraction module is almost identical to the computation of $\bm{h}_t$ in TimeGrad. However, \cite{yan2021scoregrad} have discussed the potential of adopting other network structures to encode historical information, such as temporal convolutional networks \citep{oord2016wavenet} and attention-based networks \citep{vaswani2017attention}. Here we still focus on RNN as the default choice. In the conditional SDE-based score matching module, the diffusion process is conducted through the same SDE as in Equation (\ref{eq_sde}) but its associated time-reverse SDE is refined as following:

\begin{gather}
\scalebox{1}{$\begin{aligned}
    \mathrm{d}\bm{x}_t =  \left[ f(\bm{x}_t,k) - g(k)^2 \nabla_{\bm{x}_t} \log{q_k(\bm{x}_t|\bm{h}_t)} \right]\mathrm{d}k + g(k)\mathrm{d}\bm{w},
\end{aligned}$}
\end{gather}
where $k \in [0,K]$ represents the SDE integral time. As a common practice, the conditional score function $\nabla_{\bm{x}_t} \log{q_k(\bm{x}_t|\bm{h}_t)}$ is approximated with a parameterized neural network $\bm{s}_{\bm{\theta}} (\bm{x}_t^k,\bm{h}_t,k)$. Inspired by WaveNet \citep{oord2016wavenet} and DiffWave \citep{kong2020diffwave}, the neural network is designed to have 8 connected residual blocks, while each block contains a bidirectional dilated convolution module, a gated activation unit, a skip-connection process, and an 1D convolutional neural network for output. 

The objective function of ScoreGrad is a conditional modification of Equation (\ref{eq_sdeloss}), computed as
\begin{equation}
    \sum_{t=t_0}^T L_t(\bm{\theta})
\end{equation}
with $L_t(\bm{\theta})$ being
\begin{gather}
\scalebox{1}{$\begin{aligned}
    \E_{k,\bm{x}^0_t,\bm{x}^k_t}\left[ \delta(k) \left\| \bm{s}_{\bm{\theta}} (\bm{x}_t^k,\bm{h}_t,k) - \nabla_{\bm{x}_t} \log{q_{0k}(\bm{x}_t|\bm{x}_t^0)} \right\|^2 \right].
\end{aligned}$}
\end{gather}

Up to this point, we only use the general expression of SDE for simple illustration. In the training process, one shall decide the specific type of SDE to use. Potential options include VE SDE, VP SDE, and sub-VP SDE \citep{song2020score}. The optimization varies depending on the chosen SDE because different SDEs lead to different forward transition kernel $q(\bm{x}_t^k|\bm{x}_t^{k-1})$ and also different perturbation distribution $q_{0k}(\bm{x}_t|\bm{x}_t^0)$. Finally, for forecasting, ScoreGrad utilizes the predictor-corrector sampler as in \citep{song2020score} to sample from the time-reverse SDE.

\subsection{$\text{D}^3\text{VAE}$}
In practice, we may encounter the challenge of insufficient observations. If the historical multivariate time series were recorded based on a short period, they are prone to significant level of noises due to measurement errors, sampling variability, and randomness from other sources. To address the problem of limited and noisy time series, $\text{D}^3\text{VAE}$, proposed by \cite{li2022generative}, employs a coupled diffusion process for data augmentation, and then uses a bidirectional auto-encoder (BVAE) together with denoising score matching to clear the noise. In addition, $\text{D}^3\text{VAE}$ also considers disentangling latent variables by minimizing the overall correlation for better interpretability and stability of predictions. Moreover, the mean square error (MSE) between the prediction and actual observations in the prediction window is included in the objective function, further emphasizing the role of supervision.

Assuming that the prediction window can be generated from a set of latent variables $\bm{Z}$ that follows a Gaussian distribution $q(\bm{Z}|\bm{x}^0_{1:t_0-1})$. The conditional distribution of $\bm{Z}$ is approximated with $p_{\bm{\phi}}(\bm{Z}|\bm{x}^0_{1:t_0-1})$ where $\bm{\phi}$ denotes learnable parameters. Then, the forecasting time series $\hat{\bm{x}}_{t_0:T}$ can be generated from the estimated target distribution, given by $p_{\bm{\theta}}(\bm{x}^0_{t_0:T}|\bm{Z})$. It is not difficult to see that the prediction window is still predicted based on the context window however with latent variables $\bm{Z}$ as an intermediate. 

In the coupled diffusion process, we inject noises separately into the context window and the prediction window. Different from TimeGrad which injects noises to the observation at each time point individually, the coupled diffusion process is applied to the whole period. For the context window, the same kernel as Equation (\ref{eq_forwardproperty}) is applied such that 
\begin{equation}
    \bm{x}_{1:t_0-1}^k = \sqrt{\Tilde{\alpha}_k} \bm{x}_{1:t_0-1}^0 + \sqrt{1-\Tilde{\alpha}_k} \bm{\epsilon},
\end{equation}
where $\bm{\epsilon}$ denotes the standard Gaussian noises but with a matrix rather than a vector form.

The diffusion process is further applied to the prediction window with adjusted noise levels $\alpha^\prime_k > \alpha_k$. Let $\Tilde{\alpha}^\prime_k \coloneqq \prod_{i=1}^k {\alpha^\prime_i}$, then
\begin{equation}
    \bm{x}_{t_0:T}^k = \sqrt{\Tilde{\alpha}^\prime_k} \bm{x}_{t_0:T}^0 + \sqrt{1-\Tilde{\alpha}^\prime_k} \bm{\epsilon}.
\end{equation}
This diffusion process simultaneously augments the context window and the prediction window, thus improving the generalization ability for short time series forecasting. Besides, it is proven by \cite{li2022generative} that the uncertainty caused by the generative model and the inherent noises in the observed data can both be mitigated by the coupled diffusion process.

The backward process is accomplished with two steps. The first step is to predict $\bm{x}_{t_0:T}^k$ with a BVAE as the one used in \citep{vahdat2020nvae}, which is composed of an encoder and a decoder with multiple residual blocks and takes the disturbed context window $\bm{x}_{1:t_0-1}^k$ as input. The latent variables in $Z$ are gradually generated and fed into the model in a summation manner. The output of this process is the predicted disturbed prediction window $\hat{\bm{x}}_{t_0:T}^k$. The second step involves further cleaning of the predicted data with a denoising score matching module. More specifically, the final prediction is obtained via a single-step gradient jump \citep{saremi2019neural}:
\begin{equation*}
    \hat{\bm{x}}^0_{t_0:T} \leftarrow \hat{\bm{x}}_{t_0:T}^k - \sigma_0^2 \nabla_{\hat{\bm{x}}_{t_0:T}^k} E(\hat{\bm{x}}_{t_0:T}^k; e),
\end{equation*}
where $\sigma_0$ is prescribed and $E(\hat{\bm{x}}_{t_0:T}^k; e)$ is the energy function. 

Disentanglement of latent variables $\bm{Z}$ can efficiently enhance the model interpretability and reliability for prediction \citep{li2021learning}. It is measured by the total correlation of the random latent variables $Z$. Generally, a lower total correlation implies better disentanglement which is a signal of useful information. The computation of total correlation happens synchronously with the BVAE module. 

The objective function of $\text{D}^3\text{VAE}$ consists of four components. It can be written as
\begin{gather}
\scalebox{1}{$\begin{aligned}
    w_1 D_{KL} \left(q(\bm{x}_{t_0:T}^k) \| p_{\bm{\theta}} (\hat{\bm{x}}_{t_0:T}^k)\right) + w_2 \mathcal{L}_{DSM} + w_3 \mathcal{L}_{TC} + \mathcal{L}_{MSE},
\end{aligned}$}
\end{gather}
where $w_1, w_2, w_3$ are trade-off parameters that assign significance levels to the components. The first component $D_{KL} \left(q(\bm{x}_{t_0:T}^k) \| p_{\bm{\theta}} (\hat{\bm{x}}_{t_0:T}^k)\right)$ matches the estimated target distribution with the true distribution of the prediction window. The last three components, $\mathcal{L}_{DSM}$,$\mathcal{L}_{TC}$, and  $\mathcal{L}_{MSE}$, are corresponding to the DSM module, disentanglement of latent variables, and MSE between the prediction and the truth, respectively. The parameters $\bm{\theta}$ and $\bm{\phi}$ are learned together in the training process. Eventually, forecasting samples are generated from the learned distribution $p_{\bm{\phi}}(\bm{Z}|\bm{x}^0_{1:t_0-1})$ and $p_{\bm{\theta}}(\bm{x}^0_{t_0:T}|\bm{Z})$.

\subsection{DSPD}\label{dspd}
Multivariate time series data can be considered as a record of value changes for multiple features of an entity of interest. Data are collected from the same entity, and the measuring tools normally stay unchanged during the whole observed time period. So, assuming that the change of variables over time is smooth, the time series data can be modelled as values from an underlying continuous function \citep{bilovs2022modeling}. In this case, the context window is expressed as $\bm{X}^0_c = \{ \bm{x}(1), \bm{x}(2),...,\bm{x}(t_0-1)\}$ 
and the prediction window becomes $\bm{X}^0_p = \{\bm{x}(t_0), \bm{x}(t_0+1),...,\bm{x}(T)\}$, 
where $\bm{x}(\cdot)$ is a continuous function of the time point $t$. 

Different from traditional diffusion models, the diffusion and reverse processes are no longer applied to vector observations at each time point. Alternatively, the target of interest is the continuous function $\bm{x}(\cdot)$, which means noises will be injected and removed from a function rather than a vector. Therefore, a continuous noise function $\bm{\epsilon}(\cdot)$ should take place of the noise vector $\bm{\epsilon} \sim \mathcal{N}(\bm{0},\bm{I})$ 
This function should be both continuous and tractable such that it accounts for the correlation between measurements and enables training and sampling. These requirements are effectively satisfied by designing a Gaussian stochastic process $\bm{\epsilon}(\cdot) \sim \mathcal{GP}(\bm{0},\bm{\Sigma})$ \citep{bilovs2022modeling}.

Discrete stochastic process diffusion (DSPD) is built upon the DDPM formulation but with the stochastic process $\bm{\epsilon}(\cdot) \sim \mathcal{GP}(\bm{0},\bm{\Sigma})$. It is a delight that DSPD is only slightly different from DDPM in terms of implementation. More specifically, DSPD simply replaces the commonly applied noise $\bm{\epsilon} \sim \mathcal{N}(\bm{0},\bm{I})$ with the noise function $\bm{\epsilon}(\cdot)$ whose discretized form is $\bm{\epsilon} \sim \mathcal{N}(\bm{0},\bm{\Sigma})$. Let $\bm{X}^0$ be an observed multivariate time series in a certain period of time $\bm{T}^\prime = \{t_1^\prime, t_2^\prime, ..., t_T^\prime\}$, which means $\bm{X}^0 = \{\bm{x}(t_1^\prime), \bm{x}(t_2^\prime),...,\bm{x}(t_T^\prime)\}$. In the forward process, noises are injected through the transition kernel 
\begin{equation}
    q(\bm{X}^k|\bm{X}^0) = \mathcal{N}(\sqrt{\Tilde{\alpha}_k} \bm{X}^0, (1 - \Tilde{\alpha}_k) \bm{\Sigma})
\end{equation}
Then, the following backward transition kernel is applied to recover the original data:
\begin{equation}
    p_{\bm{\theta}} (\bm{X}^{k-1}|\bm{X}^k) = \mathcal{N}(\mu_{\bm{\theta}} (\bm{X}^k, k),(1-\alpha_k)\bm{\Sigma}).
\end{equation}
Consequently, the objective function should be changed as 
\begin{gather}\label{eq_dspd_obj}
\scalebox{1}{$\begin{aligned}
    \E_{k, \bm{X}^0, \bm{\epsilon}} \left[ \delta(k) \left\|\bm{\epsilon} - \bm{\epsilon}_{\bm{\theta}} \left(\sqrt{\Tilde{\alpha}_k} \bm{X}^0 + \sqrt{1-\Tilde{\alpha}_k}\bm{\epsilon}, k \right)\right\|^2 \right],
\end{aligned}$}
\end{gather}
where $\bm{\epsilon} \sim \mathcal{N}(\bm{0},\bm{\Sigma})$ with the covariance matrix from the Gaussian process $\mathcal{GP}(\bm{0},\bm{\Sigma})$.

Forecasting via DSPD is very similar to TimeGrad. 
As before, the aim is still to learn the conditional probability $q(\bm{X}^0_p|\bm{X}^0_c)$. But there are two major improvements. Firstly, the prediction is available for any future time point in the continuous time interval. Secondly, instead of step-by-step forecasting, DSPD can generate samples for multiple time points in one run. By adding the historical condition into the fundamental objective function in Equation (\ref{eq_dspd_obj}), the objective function for DSPD forecasting is then given by
\begin{gather}
\scalebox{1}{$\begin{aligned}
    \E_{k, \bm{X}_p^0, \bm{\epsilon}} \left[ \delta(k) \left\|\bm{\epsilon} - \bm{\epsilon}_{\bm{\theta}} \left(\sqrt{\Tilde{\alpha}_k} \bm{X}_p^0 + \sqrt{1-\Tilde{\alpha}_k}\bm{\epsilon}, \bm{X}_c^0, k \right)\right\|^2 \right].
\end{aligned}$}
\end{gather}
Finally, supposing that the last context window is $\Tilde{\bm{X}}_c$, the sampling process to forecast for the prediction target $\Tilde{\bm{X}}_p$ in time interval $\bm{\Tilde{T}}$ is given by
\begin{gather*}
\scalebox{1}{$\begin{aligned}
     \Tilde{\bm{X}}_p^k \leftarrow \frac{\left( \Tilde{\bm{X}}_p^{k+1}-\zeta(k+1) \bm{L} \bm{\epsilon}_{\bm{\theta}} ( \Tilde{\bm{X}}_p^{k+1}, \Tilde{\bm{X}}_c, k+1) \right) }{\sqrt{\alpha_{k+1}}} + (1-\alpha_{k+1}) \bm{z},
\end{aligned}$}
\end{gather*}
where $\bm{L}$ is from the factorization of the covariance matrix $\bm{\Sigma} = \bm{L}\bm{L}^\intercal$, the last diffusion output is generated as $\Tilde{\bm{X}}_p^K \sim \mathcal{N} (\bm{0},\bm{\Sigma})$, and $\bm{z} \sim \mathcal{N} (\bm{0},\bm{\Sigma})$.

Similar to the extension from TimeGrad to ScoreGrad, the continuous noise function can also be adapted into the SDE framework, thus leading to the continuous stochastic process diffusion (CSPD) model \citep{bilovs2022modeling}. The diffusion process of CSPD introduces the factorized covariance matrix $\bm{\Sigma}=\bm{L}\bm{L}^\intercal$ to VP SDE (see section \ref{sde}) as
\begin{equation}
    \mathrm{d}\bm{X} = -\frac{1}{2} \alpha(k) \bm{X} \mathrm{d}k + \sqrt{\alpha (k)} \bm{L} \mathrm{d} \bm{w},
\end{equation}
where $\bm{w}$ is a matrix that represents a standard Wiener process. The perturbation distribution in the objective function is then modified as
\begin{gather}
\scalebox{1}{$\begin{aligned}
    q_{0k}(\bm{X}^k|\bm{X}^0) = \mathcal{N}\left( \bm{X}^k; \bm{X}^0e^{-\frac{1}{2} \int_0^k \alpha(s) \mathrm{d}s}, [1-e^{-\int_0^k \alpha(s)\mathrm{d}s}] \bm{\Sigma} \right).
\end{aligned}$}
\end{gather}

\subsection{DiffSTG}\label{diffstg}
Spatio-temporal graphs (STGs) are a special type of multivariate time series that encodes spatial and temporal relationships and interactions among different entities in a graph structure \citep{wen2023diffstg}. They are commonly observed in real-life applications such as traffic flow prediction \citep{li2017diffusion}, weather forecasting \citep{simeunovic2021spatio}, and finance prediction \citep{Zhou11}. Suppose we have $N$ entities of interest, such as traffic sensors or companies in the stock market. We can model these entities and their underlying relationships as a graph $\mathcal{G} = \{\mathcal{V}, \mathcal{E}, \bm{W}\}$, where $\mathcal{V}$ is a set of $N$ nodes as representations for entities, $\mathcal{E}$ is a set of links that indicates the relationship between nodes, and $\bm{W}$ is a weighted adjacency matrix that describes the graph topological structure. Multivariate time series observed at all entities are models as graph signals $\bm{X}^0_c = \{\bm{x}^0_1, \bm{x}^0_2,...,\bm{x}^0_{t_0-1} |\bm{x}^0_{t} \in \R^{D \times N}\}$, which means we have $D$-dimensional observations from $N$ entities at each time point $t$. Identical to the previous problem formulation, the aim of STG forecasting is also to predict $\bm{X}^0_p = \{\bm{x}^0_{t_0}, \bm{x}^0_{t_0+1},...,\bm{x}^0_{T} |\bm{x}^0_{t} \in \R^{D \times N}\}$ based on the historical information $\bm{X}_c$. Nevertheless, except for the time dependency on historical observations, we also need to consider the spatial interactions between different entities represented by the graph topology. 

DiffSTG applies diffusion models on STG forecasting with a graph-based noise-matching network called UGnet \citep{wen2023diffstg}. The idea of DiffSTG can be regarded as the extension of DDPM-based forecasting to STGs with an additional condition on the graph structure, which means the target distribution in Equation (\ref{eq_forecasting_dist}) is approximated alternatively by 
\begin{equation} \label{eq_diffstg_apx}
    p_{\bm{\theta}} (\bm{x}_{t_0:T}^0|\bm{x}_{1:t_0-1}^0,\bm{W}).
\end{equation}
Accordingly, the objective function is changed as
\begin{gather}
\scalebox{1}{$\begin{aligned}\label{eq_stg_obj}
    \E_{k, \bm{x}_{t_0:T}^0, \bm{\epsilon}} \left[ \delta(k) \left\|\bm{\epsilon} - \bm{\epsilon}_{\bm{\theta}} \left(\sqrt{\Tilde{\alpha}_k} \bm{x}_{t_0:T}^0 + \sqrt{1-\Tilde{\alpha}_k}\bm{\epsilon}, \bm{x}_{1:t_0-1}^0 , k, \bm{W}\right)\right\|^2 \right].
\end{aligned}$}
\end{gather}
The objective function in Equation (\ref{eq_stg_obj}) actually treats the context window and the prediction window as samples from two separate sample spaces, namely, $\bm{X}^0_c \in \mathcal{X}_c$ and $\bm{X}^0_p \in \mathcal{X}_p$ with $\mathcal{X}_c$ and $\mathcal{X}_p$ being two individual sample spaces. However, considering the fact that the context and prediction intervals are consecutive, it may be more reasonable to treat the two windows as a complete sample from the same sample space. To this end, \cite{wen2023diffstg} reformulate the forecasting problem and revise the approximation in Equation (\ref{eq_diffstg_apx}) as
\begin{equation}
    p_{\bm{\theta}} (\bm{x}_{1:T}^0|\bm{x}_{1:t_0-1}^0,\bm{W}),
\end{equation}
in which the history condition is derived by masking the future time series from the whole time period. The associated objective function is
\begin{gather}
\scalebox{1}{$\begin{aligned}
    \E_{k, \bm{x}_{1:T}^0, \bm{\epsilon}} \left[ \delta(k) \left\|\bm{\epsilon} - \bm{\epsilon}_{\bm{\theta}} \left(\sqrt{\Tilde{\alpha}_k} \bm{x}_{1:T}^0 + \sqrt{1-\Tilde{\alpha}_k}\bm{\epsilon}, \bm{x}_{1:t_0-1}^0, k,\bm{W}\right)\right\|^2 \right].
\end{aligned}$}
\end{gather}
The training process is quite straightforward following the common practice. But it should be noted that the sample generated in the forecasting process includes both historical and future values. So, we need to take out the forecasting target in the sample as the prediction. 

Now, there is only one remaining problem. How to encode the graph structural information in the noise-matching network $\bm{\epsilon}_{\bm{\theta}}$? \cite{wen2023diffstg} proposed UGnet, an Unet-based network architecture \citep{ronneberger2015u} combined with a graph neural network (GNN) to process time dependency and spatial relationships simultaneously. UGnet takes $\bm{x}_{1:T}^k, \bm{x}_{1:t_0-1}^0, k$ and $\bm{W}$ as inputs and then outputs the prediction of the associated error $\bm{\epsilon}$.


\subsection{GCRDD}
Graph convolutional recurrent denoising diffusion model (GCRDD) is another diffusion-based model for STG forecasting \citep{ruikun2023}. It differs from DiffSTG that it uses hidden states from a recurrent component to store historical information as TimeGrad and employs a different network structure for the noise-matching term $\bm{\epsilon}_{\theta}$. Please note that the notations related to STGs here follow subsection \ref{diffstg}.

GCRDD approximates the target distribution with a probabilistic density function conditional on the hidden states and graph structure as following:
\begin{equation} 
    \prod_{t=t_0}^T p_{\bm{\theta}} (\bm{x}_{t}^0|h_{t-1},\bm{W}),
\end{equation}
where the hidden state is computed with a graph-modified GRU, written as
\begin{equation}
    \bm{h}_t = \mathrm{GraphGRU}_{\bm{\theta}} (\bm{x}_t^0, \bm{h}_{t-1}, \bm{W}).
\end{equation}
The graph-modified GRU replaces the weight matrix multiplication in traditional GRU \citep{chung2014empirical} with graph convolution such that both temporal and spatial information is stored in the hidden state. The objective function of GCRDD adopts a similar form of TimeGrad but with additional graph structural information in the noise-matching network:
\begin{gather} 
\scalebox{1}{$\begin{aligned} 
    \E_{k, \bm{x}_t^0, \bm{\epsilon}} \left[ \delta(k) \left\|\bm{\epsilon} - \bm{\epsilon}_{\bm{\theta}} \left(\sqrt{\Tilde{\alpha}_k} \bm{x}_t^0 + \sqrt{1-\Tilde{\alpha}_k}\bm{\epsilon},\bm{h}_{t-1}, \bm{W}, k \right)\right\|^2 \right].
\end{aligned}$}
\end{gather}
For the noise-matching term, GCRDD adopts a variant of DiffWave \citep{kong2020diffwave} that incorporates a graph convolution component to process spatial information in $\bm{W}$. STG forecasting via GCRDD is the same as TimeGrad except that the sample generated at each time point is a matrix rather than a vector.

\section{Time Series Imputation} \label{ts_imputation}
In real-world problem settings, we usually encounter the challenge of missing values. When collecting time series data, the collection conditions may change over time, which makes it difficult to ensure the completeness of observation. In addition, accidents such as sensor failures and human errors may also result in the missing of historical records. Missing values in time series data normally have a negative impact on the accuracy of analysis and forecasting since the lack of partial observations makes the inference and conclusions vulnerable in future generalization.  

Time series imputation aims to fill in the missing values in incomplete time series data. Many previous studies have focused on designing deep learning-based algorithms for time series imputation \citep{osman2018survey}. Most existing approaches involve the RNN architecture to encode time-dependency in the imputation task \citep{che2018recurrent,cao2018brits,luo2018multivariate,yoon2018estimating}. Except for these deterministic methods, probabilistic imputation models such as GP-VAE \citep{fortuin2020gp} and V-RIN \citep{mulyadi2021uncertainty} have also shown their practical value in recent years. As a rising star in probabilistic models, diffusion models have also been applied to time series imputation tasks \citep{tashiro2021csdi, lopezalcaraz2023diffusionbased}. Compared with other probabilistic approaches, diffusion-based imputation enjoys high flexibility in the assumption of the true data distribution. In this section, we will cover four diffusion-based methods, including three for multivariate time series imputation and one for STG imputation.  

\subsection{Problem Formulation} \label{imp_problem}
We still consider the multivariate time series $\bm{X}^0=\{\bm{x}^0_1, \bm{x}^0_2,...,\bm{x}^0_T|\bm{x}^0_i \in \R^D\}$. It is not difficult to see that $\bm{X}^0 \in \R^{D \times T}$, where $D$ is the number of features, and $T$ is the number of time points in the period $[1,T]$. Different from time series forecasting in which we assume that all elements in $\bm{X}^0$ are known, here we have an incomplete matrix of observations. In the imputation task, we try to predict the values of missing data by exploring the information from some observed data. We denote the observed data as $\bm{X}^0_{ob}$ and the missing data as $\bm{X}^0_{ms}$. Then, the imputation task is to find the conditional probability distribution $q(\bm{X}^0_{ms}|\bm{X}^0_{ob})$. 

For practical purposes, zero padding is applied to the incomplete matrix $\bm{X}^0$ such that all missing entries are assigned to 0. In addition, a zero-one matrix $\bm{M} \in \R^{D\times T}$ is constructed as a mask to denote the position of missing values. More specifically, the elements in $\bm{M}$ is $0$ when the corresponding value in $\bm{X}^0$ is missing, and $1$ otherwise. 

Both $\bm{X}^0_{ob}$ and $\bm{X}^0_{ms}$ have the same dimension as $\bm{X}^0$. In the training process, a fraction of the actually observed data in $\bm{X}^0$ is randomly selected to be the true values of missing data, and the rest of the observed data will be the condition for prediction. A training mask $\bm{M}^\prime \in \R^{D \times T}$ is introduced to obtain $\bm{X}^0_{ob}$ and $\bm{X}^0_{ms}$. It is constructed by assigning 1 to entries that are corresponding to the remaining observed data in $\bm{X}^0$. Then, $\bm{X}^0_{ob}$ is computed as $\bm{X}^0_{ob} = \bm{M}^\prime \odot \bm{X}^0$, and $\bm{X}^0_{ms}$ is computed as $\bm{X}^0_{ms} = (\bm{M} -\bm{M}^\prime) \odot \bm{X}^0$, where $\odot$ denotes the element-wise matrix multiplication. In the forecasting process, on the other hand, all actually observed data are used as the condition, which means $\bm{X}^0_{ob} = \bm{M} \odot \bm{X}^0$.

It is worth mentioning that the problem formulation here is only a typical case. We will introduce later in subsection \ref{sssd} about another formulation that takes the whole time series matrix $\bm{X}^0$ as the target for generation. 

\subsection{CSDI}
Conditional Score-based Diffusion model for Imputation (CSDI) is the pioneering work on diffusion-based time series imputation \citep{tashiro2021csdi}. Identical to TimeGrad, the basic diffusion formulation of CSDI is also DDPM. However, as we have discussed in section \ref{timegrad}, the historical information is encoded by an RNN module in TimeGrad, which hampers the direct extension of TimeGrad to imputation tasks because the computation of hidden states may be interrupted by missing values in the context window. 

CSDI applies the diffusion and reverse processes to the matrix of missing data, $\bm{X}^0_{ms}$. Correspondingly, the reverse transition kernel is refined as a probabilistic distribution conditional on $\bm{X}^0_{ob}$:
\begin{gather} 
\scalebox{1}{$\begin{aligned} \label{eq_csdi_reverse_kernel}
    & p_{\bm{\theta}}(\bm{X}_{ms}^{k-1}| \bm{X}_{ms}^k, \bm{X}_{ob}^0) \\ & = \mathcal{N}(\bm{X}_{ms}^{k-1}; \bm{\mu}_{\bm{\theta}}( \bm{X}_{ms}^k,k|\bm{X}_{ob}^0),\sigma_{\bm{\theta}}( \bm{X}_{ms}^k,k|\bm{X}_{ob}^0)\bm{I}),
\end{aligned}$}
\end{gather}
where
\begin{gather}
\scalebox{1}{$\begin{aligned} \label{eq_csdi_mean}
    \bm{\mu}_{\bm{\theta}}(\bm{X}_{ms}^k,k|\bm{X}_{ob}^0) = \frac{1}{\sqrt{\alpha_k}} \left( \bm{X}_{ms}^k - \zeta(k) \bm{\epsilon}_{\bm{\theta}} (\bm{X}_{ms}^k,k|\bm{X}_{ob}^0) \right)
\end{aligned}$}
\end{gather}
with $\bm{X}_{ms}^k = \sqrt{\Tilde{\alpha}_k} \bm{X}_{ms}^0 + \sqrt{1-\Tilde{\alpha}_k}\bm{\epsilon}$. One may notice that the variance term here is different from the version in DDPM \citep{ho2020denoising}. Previously, the variance term is defined with some pre-specified constant $\sigma_k$ with $k = 1,2,...,K$, implying that the variance is treated as a hyperparameter. CSDI, however, defines a learnable version $\sigma_{\bm{\theta}}$ with parameter $\bm{\theta}$. Both ways are acceptable and have their respective practical value.

The objective function of CSDI is given by
\begin{gather}
\scalebox{1}{$\begin{aligned}
    \E_{k, \bm{X}_{ms}^0, \bm{\epsilon}} \left[ \delta(k) \left\|\bm{\epsilon} - \bm{\epsilon}_{\bm{\theta}} \left(\sqrt{\Tilde{\alpha}_k} \bm{X}_{ms}^0 + \sqrt{1-\Tilde{\alpha}_k}\bm{\epsilon}, \bm{X}_{ob}^0, k \right)\right\|^2 \right].
\end{aligned}$}
\end{gather}
The noise-matching network $\bm{\epsilon}_{\bm{\theta}}$ adopts the DiffWave \citep{kong2020diffwave} architecture by default. After training, the imputation is accomplished by generating the target matrix of missing values in the same way as DDPM. $\bm{X}_{ob}^0$ in the sampling process is identical to the zero padding version of the original time series matrix $\bm{X}^0$, where all missing values are assigned to $0$. The starting point of the sampling process is a random Gaussian imputation target $\bm{X}_{ms}^K \sim \mathcal{N}(\bm{0},\bm{I})$. Then, for $k = K-1,...,1$, the algorithm computes:
\begin{gather*}
\scalebox{1}{$\begin{aligned}
    \bm{X}_{ms}^k \leftarrow \frac{\left( \bm{X}_{ms}^{k+1} -\zeta(k+1)\bm{\epsilon}_{\bm{\theta}} (\bm{X}_{ms}^{k+1}, \bm{X}_{ob}^0, k+1) \right) }{\sqrt{\alpha_{k+1}}} + \sigma_{\bm{\theta}} \bm{Z},
\end{aligned}$}
\end{gather*} 
where $\bm{Z} \sim \mathcal{N}(\bm{0},\bm{I})$ for $k = K-1, ..., 1$, and $\bm{Z} = \bm{0}$ for $k = 0$.

\subsection{DSPD}
Here we discuss the simple extension of DSPD and CSPD in section \ref{dspd}  to time series imputation tasks. Since the rationale behind the extension of these two models is almost identical, we only focus on DSPD for illustration. The assumption of DSPD states that the observed time series is formed by values of a continuous function $\bm{x}(\cdot)$ of time $t$. Therefore, the missing values can be obtained by computing the values of this continuous function at the corresponding time points. Recall that DSPD utilizes the covariance matrix $\bm{\Sigma}$ instead of the DDPM variance $\sigma_k\bm{I}$ or $\sigma_{\bm{\theta}}$ in the backward process. Therefore, one may apply DSPD to imputation tasks in a similar way as CSDI by replacing the variance term in Equation (\ref{eq_csdi_reverse_kernel}) with the covariance matrix $\bm{\Sigma}$. According to \cite{bilovs2022modeling}, the continuous noise process is a more natural choice than the discrete noise vector because it takes account of the irregularity in the measurement when collecting the time series data. 

\subsection{SSSD} \label{sssd}
Structured state space diffusion (SSSD) differs from the aforementioned two methods by having the whole time series matrix $\bm{X}^0$ as the generative target in its diffusion module \citep{lopezalcaraz2023diffusionbased}. The name, ``structured state space diffusion'', comes from the design of the noise-matching network $\bm{\epsilon}_{\bm{\theta}}$, which adopts the state space model \citep{guefficiently} as the internal architecture. As a matter of fact, $\bm{\epsilon}_{\bm{\theta}}$ can also take other architectures such as the DiffWave-based network in CSDI \citep{tashiro2021csdi} and SaShiMi, a generative model for sequential data \citep{goel2022s}. However, the authors of SSSD have shown empirically that the structured state space model generally generates the best imputation outcome compared with other architectures \citep{lopezalcaraz2023diffusionbased}. To emphasize the difference between this method with other diffusion-based approaches, here we will primarily focus on the unique problem formulation used by SSSD. 

As we have mentioned, the generative target of SSSD is the whole time series matrix, $\bm{X}^0 \in \R^{D \times T}$, rather than a matrix that particularly represents the missing values. For the purpose of training, $\bm{X}^0$ is also processed with zero padding. The conditional information, in this case, is from a concatenated matrix $\bm{X}^0_c = \mathrm{Concat}(\bm{X}^0 \odot \bm{M}_c, \bm{M}_c)$, where $\bm{M}_c$ is a zero-one matrix indicating the position of observed values as the condition. The element in $\bm{M}_c$ can only be $1$ if its corresponding value in $\bm{X}^0$ is known. 

There are two options for the objective function used in the training process. Similar to other approaches, the objective function can be a simple conditional variant of the DDPM objective function:
\begin{gather}
\scalebox{1}{$\begin{aligned}
    \E_{k, \bm{X}^0, \bm{\epsilon}} \left[ \delta(k) \left\|\bm{\epsilon} - \bm{\epsilon}_{\bm{\theta}} \left(\sqrt{\Tilde{\alpha}_k} \bm{X}^0 + \sqrt{1-\Tilde{\alpha}_k}\bm{\epsilon}, \bm{X}_{c}^0, k \right)\right\|^2 \right],
\end{aligned}$}
\end{gather}
where $\bm{\epsilon}_{\bm{\theta}}$ is by default built upon the structured state space model. The other choice of the objective function is computed with only known data, which is mathematically expressed as
\begin{gather}
\scalebox{1}{$\begin{aligned}
    \E_{k, \bm{X}^0, \bm{\epsilon}} \left[ \delta(k) \left\|\bm{\epsilon} \odot \bm{M}_c - \bm{\epsilon}_{\bm{\theta}} \left(\sqrt{\Tilde{\alpha}_k} \bm{X}^0 + \sqrt{1-\Tilde{\alpha}_k}\bm{\epsilon},  \bm{X}_{c}^0, k \right) \odot \bm{M}_c \right\|^2 \right].
\end{aligned}$}
\end{gather}
According to \cite{lopezalcaraz2023diffusionbased}, the second objective function is typically a better choice in practice. For forecasting, SSSD employs the usual sampling algorithm and applies to the unknown entries in $\bm{X}^0$, namely, $(1-\bm{M}_c) \odot \bm{X}^0$.

An interesting point proposed along with SSSD is that imputation models can also be applied to forecasting tasks. This is because future time series can be viewed as a long block of missing values on the right of $\bm{X}^0$. Nevertheless, experiments of SSSD have shown that the diffusion-based approaches underperform other methods such as the Autoformer \citep{wu2021autoformer} in forecasting tasks.

\subsection{PriSTI}
PriSTI is a diffusion-based model for STG imputation \citep{liu2023pristi}. However, different from DiffSTG, the existing framework of PriSTI is designed for STGs with only one feature, which means the graph signal has the form $\bm{X}^0 = \{ \bm{x}^0_1,\bm{x}^0_2,...,\bm{x}^0_T\} \in \R^{N \times T}$. Each vector $\bm{x}^0_t \in \R^{N}$ represents the observed values of $N$ nodes at time point $t$. This kind of data is often observed in traffic prediction \citep{li2017diffusion} and weather forecasting \citep{yi2016st}. \textbf{\textit{METR-LA}}, for example, is an STG dataset that contains traffic speed collected by 207 sensors on a Los Angeles highway in a 4-month time period \citep{li2017diffusion}. There is only one node attribute, that is, traffic speed. However, unlike the multivariate time series matrix, where features (sensors in this case) are usually assumed to be uncorrelated, the geographic relationship between different sensors is stored in the weighted adjacency matrix $\bm{W}$, allowing a more pertinent representation of real-world traffic data. 

The number of nodes $N$ can be considered as the number of features $D$ in CSDI. The only difference is that PriSTI incorporates the underlying relationship between each pair of nodes in the conditional information for imputation. So, the problem formulation adopted by PriSTI is the same as our discussion in subsection \ref{imp_problem}, thus the goal is still to find $q(\bm{X}^0_{ms}|\bm{X}^0_{ob})$. 

To encode graph structural information, the mean in Equation (\ref{eq_csdi_mean}) is modified as
\begin{gather}
\scalebox{1}{$\begin{aligned}
    \bm{\mu}_{\bm{\theta}}(\bm{X}_{ms}^k,k|\bm{X}_{ob}^0, \bm{W}) = \frac{1}{\sqrt{\alpha_k}} \left( \bm{X}_{ms}^k - \zeta(k) \bm{\epsilon}_{\bm{\theta}} (\bm{X}_{ms}^k,\bm{X}_{ob}^0,k,\bm{W}) \right),
\end{aligned}$}
\end{gather}
where $\bm{W}$ is the weighted adjacency matrix. Consequently, the objective function is changed as
\begin{gather}
\scalebox{1}{$\begin{aligned}
    \E_{k, \bm{X}_{ms}^0, \bm{\epsilon}} \left[ \delta(k) \left\|\bm{\epsilon} - \bm{\epsilon}_{\bm{\theta}} \left(\sqrt{\Tilde{\alpha}_k} \bm{X}_{ms}^0 + \sqrt{1-\Tilde{\alpha}_k}\bm{\epsilon}, \bm{X}_{ob}^0, k, \bm{W} \right)\right\|^2 \right].
\end{aligned}$}
\end{gather}
The conditional information, $\bm{X}_{ob}^0$, is processed with linear interpolation before it is fed into the algorithm to incorporate extra noises, which will enhance the denoising capability of the model and eventually lead to better consistency in prediction \citep{choi2022graph}. The noise-matching network $\bm{\epsilon}_{\bm{\theta}}$ is composed of two modules, including a conditional feature extraction module and a noise estimation module. The conditional feature extraction module takes the interpolated information $\bm{X}_{ob}^0$ and adjacency matrix $\bm{W}$ as inputs and generates a global context with both spatial and temporal information as the condition for diffusion. Then, the noise estimation module utilizes this global context to estimate the injected noises with a specialized attention mechanism to capture temporal dependencies and geographic information. Ultimately, the STG imputation is fulfilled with the usual sampling process of DDPM, but with the specially designed noise-matching network here to incorporate the additional spatial relationship. 

Since PriSTI only works for the imputation of STGs with a single feature, which is simply a special case of STGs, this model's practical value is somehow limited. So, the extension of the idea here to more generalized STGs is a notable topic for future researchers.

\section{Time Series Generation} \label{ts_generation}
The rapid development of the machine learning paradigm requires high-quality data for different learning tasks in finance, economics, physics, and other fields. The performance of machine learning model and algorithm may highly subject to the underlying data quality. Time series generation refers to the process of creating synthetic data that resembles the real-world time series. Since the time series data is characterized by its temporal dependencies, the generation process usually requires the learning of underlying patterns and trends, from the past information. 

Time series generation is a developing topic in the literature with the existence of several methods \citep{yoon2018estimating,desai2021timevae}. Time series data can be seen as a case of sequential data, whose generation usually involves the GAN architecture \citep{xu2020cot, donahue2018adversarial,esteban2017real, mogren2016c}. Accordingly, TimeGAN is proposed to generate time series data based on an integration of RNN and GAN for the purpose of processing time dependency and generation \citep{yoon2018estimating}. However, the GAN-based generative methods have been criticized as they are unstable \citep{chu2020smoothness} and subject to the model collapse issue \citep{xiao2021tackling}. Another way to generate time series data is stemmed from the variational autoencoder, leading to the so-called TimeVAE model \citep{desai2021timevae}. As a common shortcoming of VAE-based models, TimeVAE requires a user-defined distribution for its probabilistic process. Here we will present a different probabilistic time series generator originated from diffusion models, which is more flexible with the form of the target distribution. We will particularly focus on \citep{lim2023regular} because it is the first and only work on this novel design. This section aims to enlighten researchers about this new-born research direction, and we expect to see more derivative works in the future.

\subsection{Problem Formulation}

With the multivariate time series $\bm{X}^0=\{\bm{x}^0_1, \bm{x}^0_2,...,\bm{x}^0_T|\bm{x}^0_i \in \R^D\}$, the time series generation problem aims to synthesize time series $\bm{x}^0_{1:T}$ by generating observation $\bm{x}^0_{t}$ at time point $t\in [2,T]$ with the consideration of its previous historical data $\bm{x}^0_{1:t-1}$. Correspondingly, the target distribution is the conditional density $q(\bm{x}^0_t|\bm{x}^0_{1:t-1})$ for $t \in [2,T]$, and the associated generative process involves the recursive sampling of $\bm{x}_t$ for all time points in the observed period. Details about the training and generation processes will be discussed in the next subsection.

\subsection{TSGM}
To our best knowledge, \citep{lim2023regular} is the only work to study the time series generation problem based on the diffusion method. The conditional score-based time series generative model (TSGM) was proposed, to conditionally generate each time series observation based on the past generated observations. The TSGM architecture includes three components: an encoder, a decoder and a conditional score-matching network.
The pre-trained encoder is used to embed the underlying time series into a latent space. The conditional score-matching network is used to sample the hidden states, which are then converted to the time series samples via the decoder.

Given the multivariate time series $\bm{X}^0=\{\bm{x}^0_1, \bm{x}^0_2,...,\bm{x}^0_T|\bm{x}^0_i \in \R^D\}$, the encoder $\bm{\mathrm{En}}$ and decoder $\bm{\mathrm{De}}$ enable the mapping between the time series data and hidden states in a latent space. The mapping process can be defined as:
\begin{equation}
    \bm{h}_t = \bm{\mathrm{En}}(\bm{h}^0_{t-1},\bm{x}^0_t), ~~~\bm{\hat{x}}^0_t = \bm{\mathrm{De}}(\bm{h}^0_t),
\end{equation}
where $\bm{\hat{x}}^0_t$ refers to the reconstructed time series data at time t, after the mapping process. This is a recursive process as both the encoder $\bm{\mathrm{En}}$ and decoder $\bm{\mathrm{De}}$ are constructed with the RNN structure. The training objective function $\mathcal{L}_{ED}$ for both encoder and decoder is defined as:
\begin{equation}
    \mathcal{L}_{ED} = \E_{\bm{x}^0_{1:T}}\left[\|\bm{\hat{x}}^0_{1:T}-\bm{x}^0_{1:T}\|_2^2\right].
\end{equation}

Given the auto-dependency characteristic of the time series data, learning the conditional log-likelihood function is essential. To address this, the conditional score-matching network is designed based on the SDE formulation of diffusion models. It is worth noting that TSGM focuses on the generation of hidden states rather than producing the time series directly with the sampling process. At time step t, instead of applying the diffusion process to $\bm{x}^0_{t}$, the hidden states $\bm{h}_t^0$ is diffused to a Gaussian distribution by the following forward SDE:
\begin{equation}
    \mathrm{d}\bm{h}_t = f(k, \bm{h}_t)\mathrm{d}k + g(k)\mathrm{d}\bm{\omega}
\end{equation}
where $k\in [0,K]$ refers to the integral time. With the diffused sample $\bm{h}_{1:t}^k$, the conditional score-matching network $s_{\bm{\theta}}$ learns the gradient of the conditional log-likelihood function with the following objective function:
\begin{equation}
    \mathcal{L}_{Score} = \E_{\bf{h}_{1:T}^0,k}\sum^{T}_{t=1}\left[\mathcal{L}(t,k)\right],
\end{equation}
with
\begin{gather}
\scalebox{1}{$\begin{aligned}
    \mathcal{L}(t,k) = \E_{\bm{h}_t^k}\left[\delta(k)\|s_{\bm{\theta}} (\bm{
    h}_t^k, \bm{h}_{t-1},k) - \nabla_{\bm{h}_t}\log q_{0k}(\bm{h}_t|\bm{h}_t^0)\|^2\right].
\end{aligned}$}
\end{gather}
The network architecture of $s_{\bm{\theta}}$ is designed based on U-net \citep{ronneberger2015u}, which was adopted by the classic SDE model \citep{song2020score}.

In the training process, the encoder and decoder are pre-trained using the objective $\mathcal{L}_{ED}$. They can also be trained simultaneously with the network $s_{\bm{\theta}}$, but \cite{lim2023regular} showed that the pre-training generally led to better performance. Then, to learn the score-matching network, hidden states are firstly obtained through inputting the entire time series $\bm{x}^0_{1:T}$ into the encoder, and then fed into the training algorithm with the objective function $\mathcal{L}_{Score}$. The time series generation is achieved by sampling hidden states and then applying the decoder, where the sampling process is analogous to solving the solutions to the time-reverse SDE.

The TSGM method can achieve state-of-the-art sampling quality and diversity, compared to a range of well-developed time series generation methods. However, it is still subject to the fundamental limitation that all diffusion models may have: they are generally more computationally expensive than GANs.

\section{Conclusion} \label{conclusion}
Diffusion models, a rising star in advanced generative techniques, have shown their exceptional power in various real-world applications. In recent years, many successful attempts have been made to incorporate diffusion in time series applications to boost model performance. As a compensation for the deficiency of a methodical summary and discourse on the diffusion-based approaches for time series, we have furnished a self-contained survey of these approaches while discussing the interactions and differences among them. More specifically, we have presented six models for time series forecasting, four models for time series imputation, and one model for time series generation. Although these models have shown good performance with empirical evidence, we feel obligated to emphasize that they are usually associated with very high computational costs. In addition, since most models are constructed with a high level of theoretical background, there is still a lack of deeper discussion and exploration of the rationale behind these models. This survey is expected to serve as a starting point for new researchers in this area and also an inspiration for future directions. 

\begingroup
\setstretch{1.5}
\bibliographystyle{fitee}
\bibliography{bibsample}
\endgroup

\end{document}